\pdfoutput=1

\documentclass[11pt]{article}

\usepackage[dvipsnames]{xcolor}
\definecolor{color1}{HTML}{006EB8}
\definecolor{color2}{HTML}{009B55}
\definecolor{color3}{HTML}{00A99A}
\definecolor{color4}{HTML}{3C8031}
\definecolor{color5}{HTML}{006795}
\definecolor{color6}{HTML}{00AEB3}
\definecolor{mygray}{gray}{0.93}
\definecolor{mygreen}{HTML}{3FBC9D}
\definecolor{arsenic}{rgb}{0.23, 0.27, 0.29}

\usepackage{ACL2023}

\usepackage[utf8]{inputenc}
\usepackage[T1]{fontenc}
\usepackage{hyperref} 
\usepackage{url}
\usepackage{booktabs}
\usepackage{amsfonts}
\usepackage{nicefrac}
\usepackage{microtype}
\usepackage{wrapfig}

\usepackage{lipsum}
\usepackage{times}
\usepackage{tikz}
\usepackage{latexsym}
\usepackage{microtype}
\usepackage{graphicx}
\usepackage{placeins}
\usepackage{subfigure}
\usepackage{bbm}
\usepackage{multirow}
\usepackage{enumitem}
\usepackage{tikz}
\usepackage[export]{adjustbox}
\usepackage{amsmath}
\usepackage{amssymb}
\usepackage{mathtools}
\usepackage{amsthm}
\usepackage{nccmath}
\usepackage{algorithm}
\usepackage[font=small,labelfont=bf]{caption}
\usepackage[algo2e, ruled]{algorithm2e}
\SetKwComment{Comment}{$\triangleright$\ }{}
\SetKwProg{Init}{Initialize}{}{}
\usepackage{colortbl} 

\newcommand{\Arrow}[1]{\parbox{#1}{\tikz{\draw[->](0,0)--(#1,0);}}}

\makeatletter
\newcommand\footnoteref[1]{\protected@xdef\@thefnmark{\ref{#1}}\@footnotemark}
\makeatother

\newcommand{\modelorg}[1]{\textsc{SSNeT}}
\newcommand{\modelexc}[1]{\textsc{ExSSNeT}}
\newcommand{\method}[1]{\textsc{ExSSNeT}}

\title{Exclusive Supermask Subnetwork Training for Continual Learning}

\author{\centerline{Prateek Yadav \& Mohit Bansal} \\
\centerline{Department of Computer Science}\\
\centerline{UNC Chapel Hill}\\
\centerline{\texttt{\{praty,mbansal\}@cs.unc.edu}} \\
}

\begin{document}

\maketitle

\begin{abstract}

Continual Learning (CL) methods focus on accumulating knowledge over time while avoiding catastrophic forgetting.
Recently, \citet{supsup} proposed a CL method, SupSup, which uses a randomly initialized, fixed base network (model) and finds a \textit{supermask} for each new task that selectively keeps or removes each weight to produce a \textit{subnetwork}.
They prevent forgetting as the network weights are not being updated.
Although there is no forgetting, the performance of SupSup is sub-optimal because fixed weights restrict its representational power.
Furthermore, there is no accumulation or transfer of knowledge inside the model when new tasks are learned.
Hence, we propose \method{} (\textsc{E}xclusive \textsc{S}upermask \textsc{S}ub\textsc{Ne}twork \textsc{T}raining), that performs \textit{exclusive} and \textit{non-overlapping} subnetwork weight training. This avoids conflicting updates to the shared weights by subsequent tasks to improve performance while still preventing forgetting.
Furthermore, we propose a novel \textsc{K}NN-based \textsc{K}nowledge \textsc{T}ransfer (KKT) module that utilizes previously acquired knowledge to learn new tasks better and faster.
We demonstrate that \method{} outperforms strong previous methods on both NLP and Vision domains while preventing forgetting. Moreover, \method{} is particularly advantageous for sparse masks that activate $2$-$10$\% of the model parameters, resulting in an average improvement of $8.3$\% over SupSup. Furthermore, \method{} scales to a large number of tasks (100). Our code is available at \href{https://github.com/prateeky2806/exessnet}{https://github.com/prateeky2806/exessnet}.
\end{abstract}

\setlength{\intextsep}{2pt}
\setlength{\columnsep}{2pt}

\section{Introduction}

\label{sec:intro} 
Artificial intelligence aims to develop agents that can learn to accomplish a set of tasks. Continual Learning (CL)  \citep{ring1998child,thrun1998lifelong} is crucial for this, but when a model is sequentially trained on different tasks with different data distributions, it can lose its ability to perform well on previous tasks, a phenomenon is known as \textit{catastrophic forgetting} (CF) \citep{mccloskey1989catastrophic,zhao1996incremental,thrun1998lifelong}.
This is caused by the lack of access to data from previous tasks, as well as conflicting updates to shared model parameters when sequentially learning multiple tasks, which is called \emph{parameter interference} \citep{mccloskey1989catastrophic}.

\begin{figure*}
    \centering
    \includegraphics[width=0.9\linewidth]{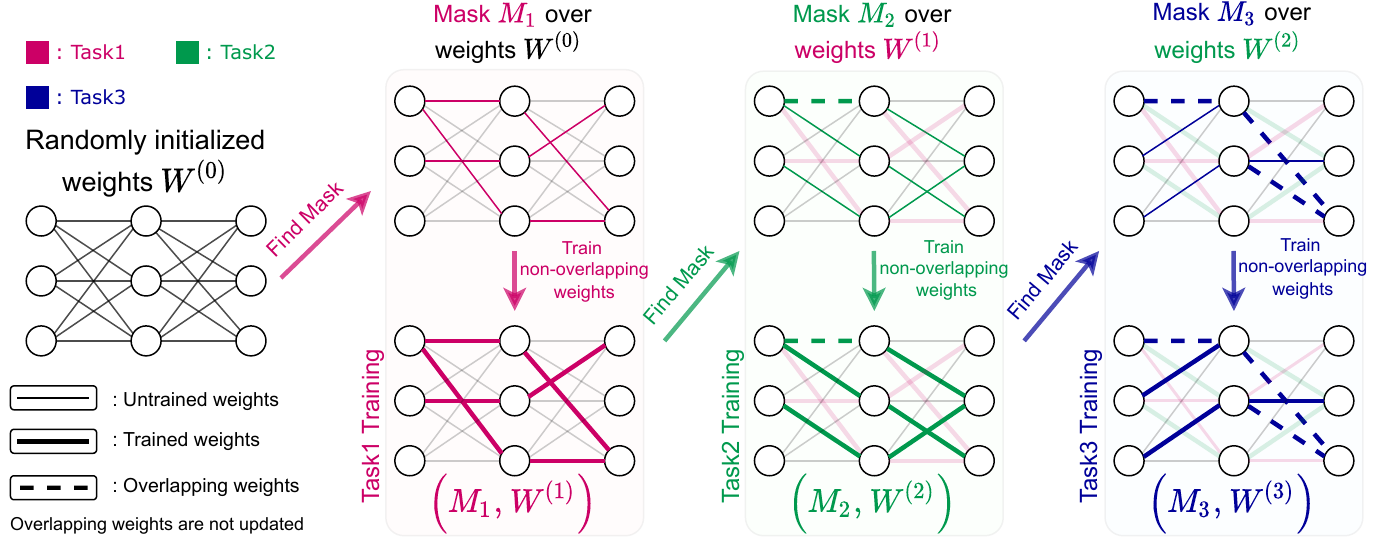}
    \captionof{figure}{\label{fig:method}\method{} diagram.
    We start with random weights $W^{(0)}$. For task 1, we first learn a supermask $M_1$ (the corresponding subnetwork is marked by red color, column 2 row 1) and then train the weight corresponding to $M_1$ resulting in weights $W^{(1)}$ (bold red lines, column 1 row 2). For task 2, we learn the mask $M_2$ over fixed weights $W^{(1)}$. If mask $M_2$ weights overlap with $M_1$ (marked by bold dashed green lines in column 3 row 1), then only the non-overlapping weights (solid green lines) of the task 2 subnetwork are updated (as shown by bold and solid green lines column 3 row 2). 
    These already trained weights (bold lines) are not updated by any subsequent task. 
    Finally, for task 3, we learn the mask $M_3$ (blue lines) and update the solid blue weights.
    }
\end{figure*}

\looseness=-1
Recently, some CL methods avoid parameter interference by taking inspiration from the \textit{Lottery Ticket Hypothesis} \citep{frankle2018lottery} and \textit{Supermasks} \citep{zhou2019deconstructing} to exploit the expressive power of sparse subnetworks.
Given that we have a combinatorial number of sparse subnetworks inside a network, \citet{zhou2019deconstructing} noted that even within randomly weighted neural networks, there exist certain subnetworks known as \textit{supermasks} that achieve good performance.
A supermask is a sparse binary mask that selectively keeps or removes each connection in a fixed and randomly initialized network to produce a subnetwork with good performance on a given task. We call this the subnetwork as \textit{supermask subnetwork} that is shown in Figure \ref{fig:method}, highlighted in red weights. 
Building upon this idea, \citet{supsup} proposed a CL method, \textit{SupSup}, which initializes a network with fixed and random weights and then learns a different supermask for each new task.
This allows them to prevent catastrophic forgetting (CF) as there is no parameter interference (because the model weights are fixed).

Although SupSup \citep{supsup} prevents CF, there are some problems with using supermasks for CL:
(1) Fixed random model weights in SupSup limits the supermask subnetwork's representational power resulting in sub-optimal performance.
(2) When learning a task, there is no mechanism for transferring learned knowledge from previous tasks to better learn the current task. Moreover, the model is not accumulating knowledge over time as the weights are not being updated.

\begin{figure}[t!]
\centering  
    \includegraphics[width=0.69\linewidth]{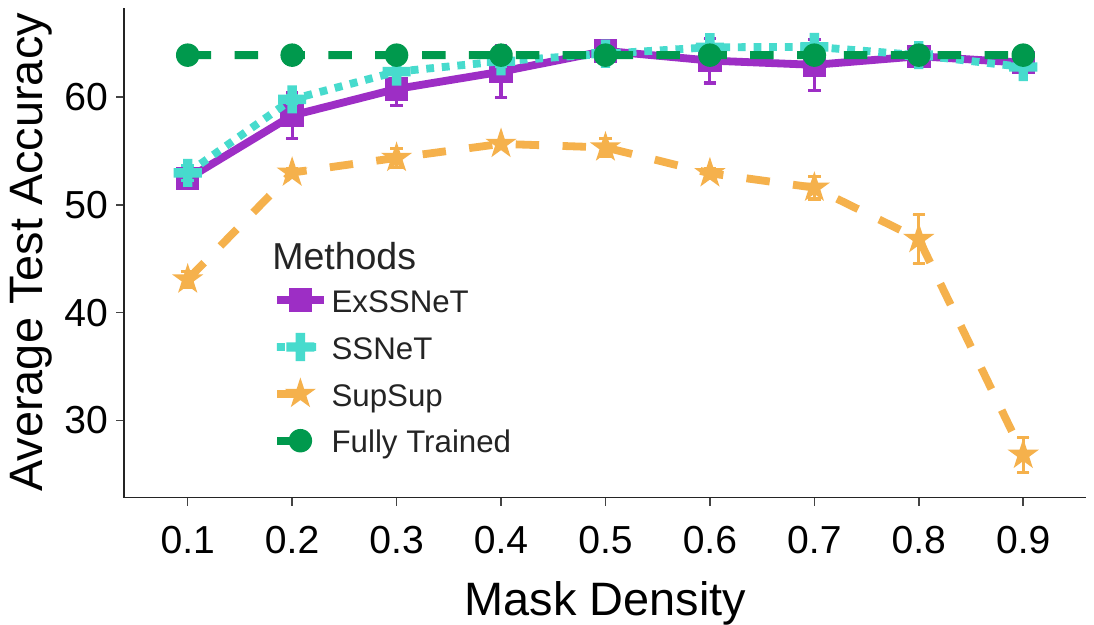}
    \vspace{-5pt}
    \captionsetup{margin=0.25cm}
    \captionof{figure}{\label{fig:single_task}Test accuracy versus the mask density for 100-way CIFAR100 classification. Averaged over 3 seeds.}
    \vspace{-10pt}
\end{figure}

\looseness=-1
To overcome the aforementioned issues, we propose our method, \method{} (\textsc{E}xclusive \textsc{S}upermask \textsc{S}ub\textsc{Ne}twork \textsc{T}raining), pronounced as \textit{`excess-net’}, which
first learns a mask for a task and then selectively trains a subset of weights from the supermask subnetwork.
We train the weights of this subnetwork via \textit{exclusion} that avoids updating parameters from the current subnetwork that have already been updated by any of the previous tasks.
In Figure \ref{fig:method}, we demonstrate \method{} that also helps us to prevent forgetting.
Training the supermask subnetwork's weights increases its representational power and allows \method{} to encode task-specific knowledge inside the subnetwork (see Figure \ref{fig:single_task}).
This solves the first problem and allows \method{} to perform comparably to a fully trained network on individual tasks; and when learning multiple tasks, the exclusive subnetwork training improves the performance of each task while still preventing forgetting (see Figure \ref{fig:sp_overlap}).

To address the second problem of knowledge transfer, we propose a $k$-nearest neighbors-based knowledge transfer (KKT) module that is able to utilize relevant information from the previously learned tasks to improve performance on new tasks while learning them faster.
Our KKT module uses KNN classification to select a subnetwork from the previously learned tasks that has a better than random predictive power for the current task and use it as a starting point to learn the new tasks.

Next, we show our method's advantage by experimenting with both natural language and vision tasks. For natural language, we evaluate on WebNLP classification tasks \citep{d2019episodic} and GLUE benchmark tasks \citep{wang-etal-2018-glue}, whereas, for vision, we evaluate on SplitMNIST \citep{zenke2017continual}, SplitCIFAR100 \citep{De_Lange_2021_cope}, and SplitTinyImageNet \citep{buzzega2020der} datasets.
We show that for both language and vision domains, \method{} outperforms multiple strong and recent continual learning methods based on replay, regularization, distillation, and parameter isolation.
For the vision domain, \method{} outperforms the strongest baseline by $4.8$\% and $1.4$\% on SplitCIFAR and SplitTinyImageNet datasets respectively, while surpassing multitask model and bridging the gap to training \textit{individual} models for each task. 
In addition, for GLUE datasets, \method{} is 2\% better than the strongest baseline methods and surpasses the performance of multitask learning that uses all the data at once.
Moreover, \method{} obtains an average improvement of $8.3$\% over SupSup for sparse masks with $2-10$\% of the model parameters and scales to a large number of tasks (100). Furthermore, \method{} with the KKT module learns new tasks in as few as 30 epochs compared to 100 epochs without it, while achieving 3.2\% higher accuracy on the SplitCIFAR100 dataset.
In summary, our contributions are listed below:
\begin{itemize}[noitemsep,topsep=0pt,leftmargin=0.5cm]
    \item We propose a simple and novel method to improve mask learning by combining it with exclusive subnetwork weight training to improve CL performance while preventing CF.
    \item We propose a \textsc{K}NN-based \textsc{K}nowledge \textsc{T}ransfer (KKT) module for supermask initialization that dynamically identifies previous tasks to transfer knowledge to learn new tasks better and faster.
    \item Extensive experiments on NLP and vision tasks show that \method{} outperforms strong baselines and is comparable to multitask model for NLP tasks while surpassing it for vision tasks. Moreover, \method{} works well for sparse masks and scales to a large number of tasks.
\end{itemize}

\vspace{-7pt}
\section{Motivation}
\vspace{-5pt}
\label{sec:motivation}

Using sparsity for CL is an effective technique to learn multiple tasks, i.e., by encoding them in different subnetworks inside a single model. 
SupSup \citep{supsup} is an instantiation of this that initializes the network weights randomly and then learns a separate supermask for each task (see Figure \ref{fig:supsup}).
They prevent CF because the weights of the network are fixed and never updated. However, this is a crucial problem as discussed below.

\paragraph{Problem 1 - Sub-Optimal Performance of Supermask:} Although fixed network weights in SupSup prevent CF, this also restricts the representational capacity, leading to worse performance compared to a fully trained network.
In Figure \ref{fig:single_task}, we report the test accuracy with respect to the fraction of network parameters selected by the mask, i.e., the \textit{mask density} for an underlying ResNet18 model on a \textit{single 100-way classification} on CIFAR100 dataset.
The fully trained ResNet18 model (dashed green line) achieves an accuracy of $63.9\%$. 
Similar to \citet{zhou2019deconstructing}, we observe that the performance of SupSup (yellow dashed line) is at least $8.3$\% worse compared to a fully trained model. 
As a possible \emph{partial} remedy, we propose a simple solution, \modelorg{} (\textsc{S}upermask \textsc{S}ub\textsc{Ne}twork \textsc{T}raining), that first finds a subnetwork for a task and then trains the subnetwork's weights. 
This increases the representational capacity of the subnetwork because there are more trainable parameters.
For a single task, the test accuracy of \modelorg{} is better than SupSup for all mask densities and matches the performance of the fully trained model beyond a density threshold. But as shown below, when learning multiple tasks sequentially, \modelorg{} gives rise to parameter interference that results in CF.

\begin{figure}[t!]
\centering
    \includegraphics[width=0.9\linewidth]{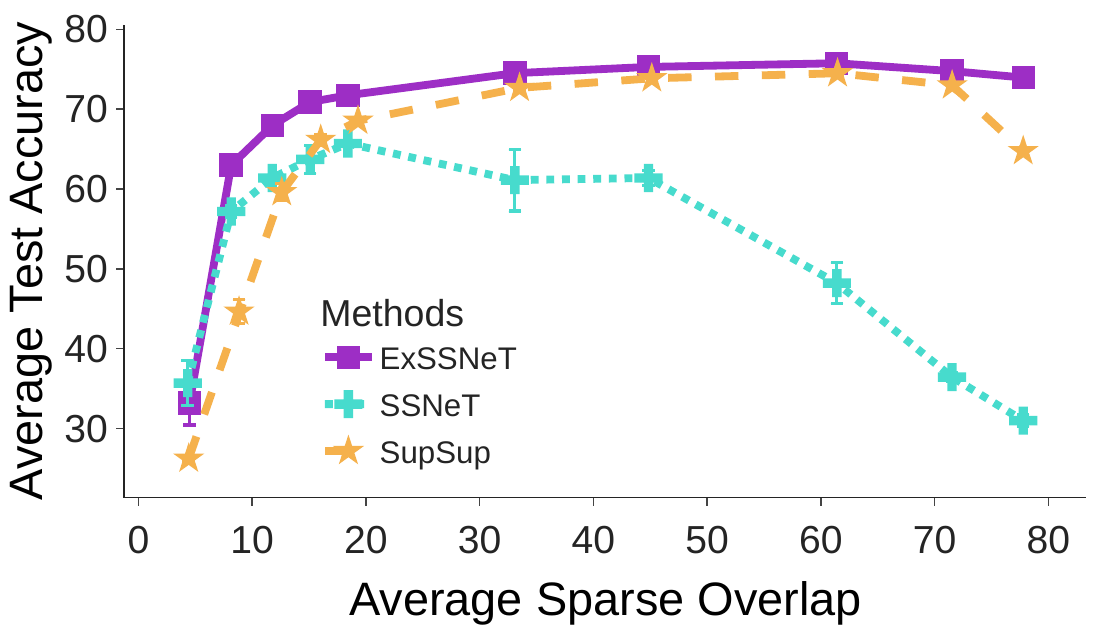}
    \captionof{figure}{\label{fig:sp_overlap}Average Test accuracy on five 20-way tasks from SplitCIFAR100  versus sparse overlap. Averaged over 3 seeds.}
\end{figure}

\paragraph{Problem 2 - Parameter Interference Due to Subnetwork Weight Training for Multiple Tasks:}
Next, we demonstrate that when learning multiple tasks sequentially, \modelorg{} can still lead to CF.
In Figure \ref{fig:sp_overlap}, we report the average test accuracy versus the fraction of overlapping parameters between the masks of different tasks, i.e., the \textit{sparse overlap} (see Equation \ref{eqn:sp_overlap}) for five different 20-way classification tasks from SplitCIFAR100 dataset with ResNet18 model.
We observe that \modelorg{} outperforms SupSup for lower sparse overlap but as the sparse overlap increases, the performance declines because the supermask subnetworks for different tasks have more overlapping (common) weights (bold dashed lines in Figure \ref{fig:method}). This leads to higher parameter interference resulting in increased forgetting which suppresses the gain from subnetwork weight training.

Our \textit{final proposal}, \method{}, resolves both of these problems by selectively training a subset of the weights in the supermask subnetwork to prevent parameter interference. When learning multiple tasks, this prevents CF, resulting in strictly better performance than SupSup (Figure \ref{fig:sp_overlap}) while having the representational power to match bridge the gap with fully trained models (Figure \ref{fig:single_task}). 

\section{Method}
\label{sec:method}

As shown in Figure \ref{fig:method}, when learning a new task $t_i$, \method{} follows three steps: (1) We learn a supermask $M_i$ for the task; 
(2) We use all the previous tasks' masks $M_1, \hdots, M_{i-1}$ to create a free parameter mask $M^{free}_i$, that finds the parameters selected by the mask $M_i$ that were not selected by any of the previous masks;
(3) We update the weights corresponding to the mask $M^{free}_i$ as this avoids parameter interference.
Now, we formally describe all the step of our method \method{} (\textsc{E}xclusive \textsc{S}upermask \textsc{S}ub\textsc{Ne}twork \textsc{T}raining) for a Multi-layer perceptron (MLP).

\paragraph{Notation:}
During training, we can treat each layer $l$ of an MLP network separately. An intermediate layer $l$ has $n_l$ nodes denoted by $\mathcal{V}^{(l)} = \{v_1, \hdots, v_{n_l}\}$. For a node $v$ in layer $l$, let $\mathcal{I}_v$ denote its input and $\mathcal{Z}_v = \sigma(\mathcal{I}_v)$ denote its output, where $\sigma(.)$ is the activation function. Given this notation, $\mathcal{I}_v$ can be written as $\mathcal{I}_v = \sum_{u \in \mathcal{V}^{(l-1)}} w_{uv}\mathcal{Z}_u$, where $w_{uv}$ is the network weight connecting node $u$ to node $v$. The complete network weights for the MLP are denoted by $W$.
When training the task $t_i$, we have access to the supermasks from all previous tasks $\{M_{j}\}_{j=1}^{i-1}$ and the model weights $W^{(i-1)}$ obtained after learning task $t_{i-1}$.

\subsection{\method{}: \textsc{E}xclusive \textsc{S}upermask \textsc{S}ub\textsc{Ne}twork \textsc{T}raining}
\label{sec:exssnet}

\paragraph{Finding Supermasks:} Following \citet{supsup}, we use the algorithm of \citet{ramanujan2019s} to learn a supermask $M_i$ for the current task $t_i$. The supermask $M_i$ is learned with respect to the underlying model weights $W^{(i-1)}$ and the mask selects a fraction of weights that lead to good performance on the task without training the weights. 
To achieve this, we learn a score $s_{uv}$ for each weight $w_{uv}$, and once trained, these scores are thresholded to obtain the mask.
Here, the input to a node $v$ is 
$\mathcal{I}_v = \sum_{u \in \mathcal{V}^{(l-1)}} w_{uv}\mathcal{Z}_u m_{uv}$,
where $m_{uv} = h(s_{uv})$ is the binary mask value and $h(.)$ is a function which outputs 1 for top-$k\%$ of the scores in the layer with $k$ being the mask density.
Next, we use a straight-through gradient estimator \citep{bengio2013straightthrough} and iterate over the current task's data samples to update the scores for the corresponding supermask $M_i$ as follows, 
\begin{equation}
    \footnotesize
    \resizebox{\linewidth}{!}{
    $
    s_{uv} = s_{uv} - \alpha \hat{g}_{s_{uv}}~;~\hat{g}_{s_{uv}} = \frac{\partial\mathcal{L}}{\partial \mathcal{I}_v}\frac{\partial \mathcal{I}_v}{\partial s_{uv}} = \frac{\partial\mathcal{L}}{\partial \mathcal{I}_v} w_{uv} \mathcal{Z}_u
    $
    }
\end{equation}
\looseness=-1

\paragraph{Finding Exclusive Mask Parameters:} 
Given a learned mask $M_i$, we use all the previous tasks' masks $M_1, \hdots, M_{i-1}$ to create a free parameter mask $M^{free}_i$, that finds the parameters selected by the mask $M_i$ that were not selected by any of the previous masks.
We do this by -- (1) creating a new mask $M_{1:i-1}$ containing all the parameters already updated by any of the previous tasks by taking a union of all the previous masks $\{M_j\}_{j=1}^{i-1}$ by using the logical \textit{or} operation, and (2) Then we obtain a mask $M^{free}_i$ by taking the intersection of all the network parameters not used by any previous task which is given by the negation of the mask $M_{1:i-1}$ with the current task mask $M_i$ via a logical \textit{and} operation. Next, we use this mask $M^{free}_i$ for the exclusive supermask subnetwork weight training.

\looseness=-1
\paragraph{Exclusive Supermask Subnetwork Weight Training:}
For training the subnetwork parameters for task $t_i$ given the free parameter mask $M^{free}_i$, we perform the forward pass on the model as $model(x, W \odot \hat{M}_i)$ where $\hat{M}_i = M^{free}_i + ((1-M^{free}_i) \odot M_i).detach()$, where $\odot$ is the element-wise multiplication.
Hence, $\hat{M}_i$ allows us to use all the connections in $M_i$ during the forward pass of the training but during the backward pass, only the parameters in $M^{free}_i$ are updated because the gradient value is 0 for all the weights $w_{uv}$ where $m^{free}_{uv} = 0$. While during the inference on task $t_i$ we use the mask $M_i$.
In contrast, \modelorg{} uses the task mask $M_i$ both during the training and inference as $model(x, W^{(i-1)} \odot M_i)$. This updates all the parameters in the mask including the parameters that are already updated by previous tasks that result in CF. Therefore, in cases where the sparse overlap is high, \method{} is preferred over \modelorg{}.
To summarize, \method{} circumvents the CF issue of \modelorg{} while benefiting from the subnetwork training to improve overall performance as shown in Figure \ref{fig:sp_overlap}.

\vspace{-2pt}
\subsection{KKT: Knn-Based Knowledge Transfer}
\label{sec:kkt_module}
\looseness=-1
When learning multiple tasks, it is a desired property to transfer information learned by the previous tasks to achieve better performance on new tasks and to learn them faster \citep{biesialska-etal-2020-continualsurvery}.
Hence, we propose a K-Nearest Neighbours (KNN) based knowledge transfer (KKT) module that uses KNN classification to dynamically find the most relevant previous task \citep{veniat2021efficient} to initialize the supermask for the current task. 
To be more specific, before learning the mask $M_i$ for the current task $t_i$, we randomly sample a small fraction of data from task $t_i$ and split it into a train and test set. Next, we use the trained subnetworks of each previous task $t_1, \hdots, t_{i-1}$ to obtain features on this sampled data. Then we learn $i-1$ independent KNN-classification models using these features. Then we evaluate these $i-1$ models on the sampled test set to obtain accuracy scores which denote the predictive power of features from each previous task for the current task. Finally, we select the previous task with the highest accuracy on the current task. If this accuracy is better than random then we use its mask to initialize the current task's supermask. This enables \method{} to transfer information from the previous task to learn new tasks better and faster.
We note that the KKT module is not limited to SupSup and can be applied to a broader category of CL methods that introduce additional parameters for new tasks.

\section{Experiments}
\vspace{-6pt}
\label{sec:experiments}

\subsection{Experimental Setup and Training Details}
\label{sec:setup}

\paragraph{Datasets:} For natural language domain, we follow the shared text classification setup of IDBR \citep{huang-etal-2021-continual-idbr}, LAMOL \citep{sun2019lamol}, and MBPA++ \citep{de2019continual} to sequentially learn five text classification tasks;
(1) Yelp Sentiment analysis \citep{zhang2015character}; 
(2) DBPedia for Wikipedia article classification \citep{dbpedia_dataset}
(3) Yahoo! Answer for Q\&A classification \citep{yahoo_dataset};
(4) Amazon sentiment analysis \citep{amazon_dataset}
(5) AG News for news classification \citep{zhang2015character}.
We call them WebNLP classification tasks for easier reference.
While comparing with the previous state-of-the-art text methods, we use the same training and test set as IDBR and LAMOL containing 115,000/500/7,600 Train/Val/Test examples. For our ablation studies, we follow IDBR and use a sampled dataset, please see Appendix Table \ref{app:tab:sampled-stats} for statistics. Additionally, we create a CL benchmark using the popular \textit{GLUE classification} tasks \citep{wang-etal-2018-glue} consisting of more than 5k train samples. We use the official validation split as test data and use $0.1\%$ of the train data to create a validation set. Our final benchmark includes five tasks; MNLI \textcolor{arsenic}{(353k/39k/9.8k)}, QQP \textcolor{arsenic}{(327k/36k/40k)}, QNLI \textcolor{arsenic}{(94k/10k/5.4k)}, SST-2 \textcolor{arsenic}{(60k/6.7k/872)}, CoLA \textcolor{arsenic}{(7.6k/856/1k)}.
For vision experiments, we follow SupSup and use three CL benchmarks, 
SplitMNIST \citep{zenke2017continual},
SplitCIFAR100 \citep{chaudhry2018efficient}, and SplitTinyImageNet \citep{buzzega2020der} datasets with 10, 100 and 200 total classes respectively.

\looseness=-1
\paragraph{Metrics:} We follow \citet{chaudhry2018efficient} and evaluate our model after learning task $t$ on all the tasks, denoted by $\mathcal{T}$. This gives us an accuracy matrix $A \in \mathbb{R}^{n \times n}$, where $a_{i,j}$ represents the classification accuracy on task $j$ after learning task $i$.
We want the model to perform well on all the tasks it has been learned. This is measured by the \textit{average accuracy}, $\mathcal{A}(\mathcal{T}) = \frac{1}{N} \sum_{k=1}^{N} a_{N,k}$, where $N$ is the number of tasks. Next, we want the model to retain performance on the previous tasks when learning multiple tasks. This is measured by the \textit{forgetting metric} \citep{lopez2017gradient}, $F(\mathcal{T}) = \frac{1}{N-1} \sum_{t=1}^{N-1} (\max_{k \in \{1, \hdots, N-1\}} a_{k,t} - a_{N,t})$. This is the average difference between the maximum accuracy obtained for task $t$ and its final accuracy. Higher accuracy and lower forgetting are desired.

\begin{table*}[!t]
\centering
\large
\resizebox{0.8\linewidth}{!}{  
\begin{tabular}{llllllllll}
\toprule
\textbf{Method ($\downarrow$)} & \multicolumn{1}{c}{\textbf{GLUE}}  & \multicolumn{5}{c}{\textbf{WebNLP}} \\
\cmidrule(lr){2-2} \cmidrule(lr){3-7}
\textbf{Order ($\rightarrow$)} & \textbf{S1} & \textbf{S2} & \textbf{S3} & \textbf{S4} & \textbf{S5} & \textbf{Average}\\

\midrule

\rowcolor{mygray}
\textbf{\textit{Random}} & \textit{33.3 (-)} & \textit{7.14 (-)} & \textit{7.14 (-)} & \textit{7.14 (-)} & \textit{7.14 (-)} & \textit{7.14 (-)} \\
\rowcolor{mygray}
\textbf{\textit{Multitask}} & \textit{79.9 (0.0)} & \textit{77.2 (0.0)} & \textit{77.2 (0.0)} & \textit{77.2 (0.0)} & \textit{77.2 (0.0)} & \textit{77.2 (0.0)} \\
\rowcolor{mygray}
\textbf{\textit{Individual}} & \textit{87.7 (0.0)} & \textit{79.5 (0.0)} & \textit{79.5 (0.0)} & \textit{79.5 (0.0)} & \textit{79.5 (0.0)} & \textit{79.5 (0.0)} \\

\midrule

\textbf{FT} & 14.1 (86.0) & 26.9 (62.1) & 22.8 (67.6) & 30.6 (55.9) & 15.6 (76.8) & 24.0 (65.6)  \\
\textbf{AdaptBERT + FT} & 24.7 (53.4) & 20.8 (68.4) & 19.1 (70.9) & 23.6 (64.5) & 14.6 (76.0) & 19.6 (70.0) \\
\textbf{AdaptBERT + Replay} & 76.8 (3.8) & 73.2 (3.0) & 74.5 (2.0) & 74.5 (2.0) & 74.6 (2.0) & 74.2 (2.3) \\
\textbf{MultiAdaptBERT} & 78.5 (0.0) & 76.7 (0.0) & 76.7 (0.0) & 76.7 (0.0) & 76.7 (0.0) & 76.7 (0.0) \\
\textbf{Prompt Tuning} & 76.3 (0.0) & 66.3 (0.0) & 66.3 (0.0) & 66.3 (0.0) & 66.3 (0.0) & 66.3 (0.0) \\
\textbf{Regularization} & 72.5 (8.8) & 76.0 (2.8) & 74.9 (3.8) & 76.4 (1.8) & 76.5 (2.0) & 76.0 (2.6) \\
\textbf{Replay} & 77.7 (4.8) & 75.1 (3.1) & 74.6 (3.5) & 75.2 (2.2) & 75.7 (3.1) & 75.1 (3.0) \\
\textbf{MBPA++}$\dagger$ & - & 74.9 (-) & 73.1 (-) & 74.9 (-) & 74.1 (-) & 74.3 (-) \\
\textbf{LAMOL}$\dagger$ & - &  76.1 (-) & 76.1 (-) & \textbf{77.2 (-)} & 76.7 (-) & 76.5 (-)  \\
\textbf{IDBR} & 73.0 (6.8) & 75.9 (2.7) & 75.4 (3.5) & 76.5 (1.6) & 76.4 (1.9) & 76.0 (2.4) \\
\textbf{SupSup} & 78.3 (0.0) & 75.9 (0.0) & 76.1 (0.0) & 76.0 (0.0) & 75.9 (0.0) & 76.0 (0.0) \\
\midrule
\textbf{\modelorg{}} & 78.4 (3.6) & 76.3 (0.8) & 76.3 (0.8) & 76.4 (0.3) & 76.3 (0.3) & 76.3 (0.6)  \\
\textbf{\method{}} & \textbf{80.5 (0.0)} & \textbf{77.0 (0.0)} & \textbf{77.1 (0.0)} & 76.7 (0.0) & \textbf{76.9 (0.0)} & \textbf{76.9 (0.0)} \\

\bottomrule
\end{tabular}
}
\vspace{-5pt}
\captionof{table}{\label{tab:text-main} Comparing average test accuracy $\uparrow$ \textcolor{BrickRed}{(\textbf{and forgetting metric} $\downarrow$}\textbf{)} for multiple tasks and sequence orders with state-of-the-art (SotA) methods. Results with $\dagger$ are taken from \citep{huang-etal-2021-continual-idbr}.
}
\vspace{-10pt}
\end{table*}

\looseness=-1
\paragraph{Sparse Overlap to Quantify Parameter Interference:}
Next, we propose \textit{sparse overlap}, a measure to quantify parameter interference for a task $i$, i.e., the fraction of the parameters in mask $M_i$ that are already updated by some previous task. 
For a formal definition refer to Appendix \ref{sec:app_spoverlap}

\paragraph{Previous Methods and Baselines:}
For both vision and language (\textbf{VL}) tasks, we compare with:
\textbf{(VL.1) Naive Training} \citep{yogatama2019learning}: where all model parameters are sequentially trained/finetuned for each task.
\textbf{(VL.2) Experience Replay (ER)} \citep{d2019episodic}: we replay previous tasks examples when we train new tasks; 
\textbf{(VL.3) Multitask Learning} \citep{Crawshaw2020MultiTaskLW}: where all the tasks are used jointly to train the model;
\textbf{(VL.4) Individual Models}: where we train a separate model for each task. This is considered an upper bound for CL;
\textbf{(VL.5) Supsup} \citep{supsup}.
For natural language (\textbf{L}), we further compare with the following methods:
\textbf{(L.6) Regularization} \citep{huang-etal-2021-continual-idbr}: Along with the Replay method, we regularize the hidden states of the BERT classifier with an L2 loss term; 
We show three Adapter BERT \citep{pmlr-v97-houlsby19adapter} variants, 
\textbf{(L.7) AdaptBERT + FT} where we have single adapter which is finetuned for all task; 
\textbf{(L.8) AdaptBERT + ER} where a single adapter is finetuned with replay; 
\textbf{(L.9) MultiAdaptBERT} where a separate adapter is finetuned for each task;
\textbf{(L.10) Prompt Tuning} \citep{li-liang-2021-prefix} that learns 50 different continuous prompt tokens for each task.
\textbf{(L.11) MBPA++} \citep{d2019episodic} perform replay with random examples during training and does local adaptation during inference to select replay example; 
\textbf{(L.12) LAMOL} \citep{sun2019lamol} uses a language model to generate pseudo-samples for previous tasks for replay; 
\textbf{(L.13) IDBR} \citep{huang-etal-2021-continual-idbr} disentangles hidden representations into generic and task-specific representations and regularizes them while also performing replay.
For vision task (\textbf{V}), we additionally compare with two popular regularization based methods, \textbf{(V.6) Online EWC} \citep{schwarz2018progress},
\textbf{(V.7) Synaptic Intelligence (SI)} \citep{zenke2017continual};
one knowledge distillation method,
\textbf{(V.8) Learning without Forgetting (LwF)} \citep{li2017learning}, 
three additional experience replay method, 
\textbf{(V.9) AGEM} \citep{chaudhry2018efficient}, 
\textbf{(V.10) Dark Experience Replay (DER)} \citep{buzzega2020der}, 
\textbf{(V.11) DER++} \citep{buzzega2020der}, and a parameter isolation method
\textbf{(V.12) CGATE} \citep{abati2020cgate}.

\paragraph{Implementation Details:} 
Following \citet{huang-etal-2021-continual-idbr}, for WebNLP datasets we learn different task orders S1-S5\footnote{\label{foot:order}For example, in S2 order the model learns the task in this order, ag \Arrow{.2cm} yelp \Arrow{.2cm} amazon \Arrow{.2cm} yahoo \Arrow{.2cm} dbpedia} that are provided in Appendix Table \ref{tab:task-order}. Following \citet{huang-etal-2021-continual-idbr}, for NLP experiments, we use a pre-trained BERT as our base model for all methods. For SupSup, \modelorg{}, and \method{}, we use a CNN-based classification head.
Unless specified, we randomly split all the vision datasets to obtain five tasks with disjoint classes. For the vision experiments, we do not use pre-trained models. 
All methods employ the same number of epochs over datasets. For additional implementation details refer to Appendix \ref{sec:additional-details}.

\subsection{Main Results} 
\label{sec:results}

\begin{table}[t!]
    \centering
    \large
    \resizebox{0.9\linewidth}{!}{  
    \begin{tabular}{cccc}
    \toprule
    \textbf{Method} & \textbf{S-MNIST} & \textbf{S-CIFAR100} & \textbf{S-TinyImageNet} \\
    \midrule
    \rowcolor{mygray}
    \textbf{\textit{Multitask}} & \textit{96.5 (0.0)} & \textit{53.0 (0.0)} & \textit{45.9 (0.0)} \\
    \rowcolor{mygray}
    \textbf{\textit{Individual}} & \textit{99.7 (0.0)} & \textit{75.5 (0.0)} & \textit{53.7 (0.0)} \\
    \midrule
    \textbf{Naive Sequential} & 49.6 (25.0) & 19.3 (73.7) & 11.5 (43.9) \\

    \textbf{EWC}	& 96.1 (4.5) & 32.4 (60.5) & 20.5 (52.1) \\
    \textbf{SI}	& 99.2 (0.6) & 46.1 (47.8) & 19.5 (46.2) \\
    \textbf{LwF}	& 99.2 (0.8) & 29.5 (70.2) & 18.1 (56.5) \\
    \textbf{AGEM} & 98.3 (1.9) & 52.1 (42.0) & 21.6 (54.9) \\
    \textbf{ER} & 99.2 (0.6) & 60.1 (27.5) & 35.6 (36.0) \\
    \textbf{DER} & 98.9 (1.2) & 62.5 (28.4) & 35.9 (37.7) \\
    \textbf{DER++} & 98.3 (1.8) & 62.5 (27.5) & 36.2 (35.7) \\
    \textbf{CGATE} & 99.6 (0.0) & 60.1 (0.0) & 49.2 (0.0) \\
    \textbf{SupSup} & 99.6 (0.0) & 62.1 (0.0) & 50.6 (0.0) \\
    \midrule
    \textbf{\modelorg{}} & \textbf{99.7 (0.0)} & 23.9 (54.4) & 49.6 (1.9) \\
    \textbf{\method{}} & \textbf{99.7 (0.0)} & \textbf{67.3 (0.0)} & \textbf{52.0 (0.0)} \\
    \bottomrule
    \end{tabular}
    }
    \vspace{-5pt}
    \captionsetup{margin=0.25cm}
    \captionof{table}{\label{tab:vision-main}Average accuracy $\uparrow$ \textcolor{BrickRed}{\textbf{(Forgetting metric} $\downarrow$\textbf{)}} on all tasks for vision. For our method, we report the results are averaged over three random seeds.}
    \vspace{-5pt}
\end{table}

\looseness=-1

\paragraph{Q1. Does Supermask Subnetwork Training Help?}

In these experiments, we show that \method{} outperforms multiple strong baseline methods including SupSup. For our main language experiments in Table \ref{tab:text-main}, we sequentially learn multiple task orders, S1 - S5\textsuperscript{\ref{foot:order}} corresponding to the GLUE and WebNLP benchmarks. These task orders are listed in Appendix Table \ref{tab:task-order}. We report the average test accuracy (and forgetting in parentheses).
For natural language, we perform better than previous SOTA CL methods in four out of five cases, across multiple task orders, and in aggregate. Specifically, on the GLUE benchmark, \method{} is at least $2.0$\% better than other methods while avoiding CF. Furthermore, \method{} either outperforms or is close to the performance of the multitasking baseline which is a strong baseline for CL methods.

\looseness=-1
For vision tasks, we split the MNIST, CIFAR100, and TinyImageNet datasets into \textit{five different tasks} with an equal number of disjoint classes and report results.
From Table \ref{tab:vision-main}, we observe that \method{} leads to a $4.8$\% and $1.4$\% improvement over the strongest baseline for Split-CIFAR100 and Split-TinyImageNet datasets.
Furthermore, both \method{} and SupSup outperform the multitask baseline. Moreover, \method{} bridges the gap to individually trained models significantly, for TinyImageNet we reach within $1.7$\% of individual models' performance. The average sparse overlap of \method{} is $19.4$\% across all three datasets implying that there is a lot more capacity in the model. See appendix Table \ref{tab:overlap} for sparse overlap of other methods and Appendix \ref{sec:app_imagenet} for best-performing methods results on Imagenet Dataset.

Note that, past methods require tricks like local adaptation in MBPA++, and experience replay in AGEM, DER, LAMOL, and ER. In contrast, \method{} is simple and does not require replay.

\begin{table}[t!]
        \centering
        \resizebox{0.9\linewidth}{!}{  
        \begin{tabular}{clll}
            
            \toprule
            \textbf{Method} & \textbf{S-MNIST} &  \textbf{S-CIFAR100} & \textbf{S-TinyImageNet} \\
            \midrule
            \textbf{SupSup} & 99.6 & 62.1 & 50.6 \\
            \textbf{+ KKT} & 99.6 [+0.0] & \textbf{67.1} [\textcolor{color2}{+5.0}] & \textbf{53.3} [\textcolor{color2}{+2.7}] \\
            \midrule
            \textbf{\modelorg{}} & \textbf{99.7} & \textbf{23.9} & 49.6 \\
            \textbf{+ KKT} & 99.3 [\textcolor{BrickRed}{-0.4}] & 23.5 [\textcolor{BrickRed}{-0.4}] & \textbf{51.8} [\textcolor{color2}{+2.2}] \\
            \midrule
            \textbf{\method{}} & 99.7 & 67.3 & 52.0 \\
            \textbf{+ KKT} & 99.7 [+0.0] & \textbf{70.5} [\textcolor{color2}{+3.2}] & \textbf{54.0} [\textcolor{color2}{+2.0}] \\
            \bottomrule
        \end{tabular}
        }
        \vspace{-5pt}
        \captionsetup{margin=0.25cm}
        \captionof{table}{\label{tab:knn-results} Average test accuracies $\uparrow$ \textcolor{color2}{\textbf{[and gains from KKT]}} when using the KKT knowledge sharing module.
        }
        
\end{table}

\begin{figure}[t!]
    \centering   
    \vspace{-10pt}
    \includegraphics[width=0.85\linewidth]{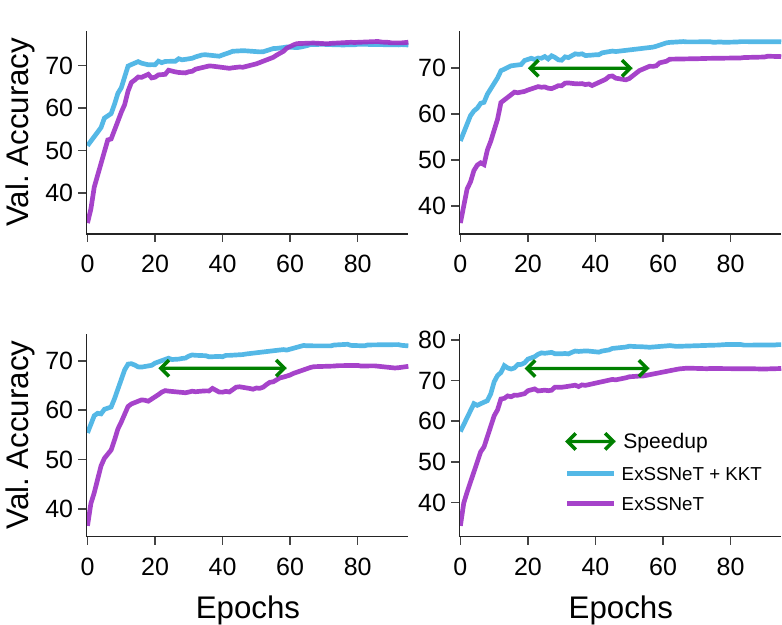}
    \vspace{-5pt}
    \captionof{figure}{We plot validation accuracy vs Epoch for \method{} and \method{} + KKT.
    We observe that KKT helps to learn the subsequent tasks faster and improves performance.
    }
    \vspace{-10pt}
    \label{fig:knn_speed}
\end{figure}

\paragraph{Q2. Can KKT Knowledge Transfer Module Share Knowledge Effectively?}

In Table \ref{tab:knn-results}, we show that adding the KKT module to \method{}, \modelorg{}, and SupSup improves performance on vision benchmarks. The experimental setting here is similar to Table \ref{tab:vision-main}. We observe across all methods and datasets that the KKT module improves average test accuracy. Specifically, for the Split-CIFAR100 dataset, the KKT module results in $5.0$\%, and $3.2$\% improvement for SupSup and \method{} respectively; while for Split-TinyImageNet, \method{} + KKT outperforms the individual models. 
We observe a performance decline for \modelorg{} when using KKT because KKT promotes sharing of parameters across tasks which can lead to worse performance for \modelorg{}.
Furthermore, \method{} + KKT outperforms all other methods on both the Split-CIFAR100 and Split-TinyImageNet datasets. 
For \method{} + KKT, the average sparse overlap is $49.6$\% across all three datasets (see appendix Table \ref{tab:overlap}). These results suggest that combining weight training with the KKT module leads to further improvements.

\paragraph{Q3. Can KKT Knowledge Transfer Module Improve Learning Speed of Subsequent Tasks?}

\looseness=-1
Next, we show that the KKT module enables us to learn new tasks faster. To demonstrate this, in Figure \ref{fig:knn_speed} we plot the running mean of the validation accuracy vs epochs for different tasks from the Split-CIFAR100 experiment in Table \ref{tab:knn-results}. We show curves for \method{} with and without the KKT module and omit the first task as both these methods are identical for Task 1 because there is no previous task to transfer knowledge.
For all the subsequent tasks (Task 2,3,4,5), we observe that -- (1) \method{} + KKT starts off with a much better initial performance compared to \method; (2) given a fixed number of epochs for training, \method{} + KKT always learns the task better because it has a better accuracy at all epochs; and (3) \method{} + KKT can achieve similar performance as \method{} in much fewer epochs as shown by the green horizontal arrows.
This clearly illustrates that using the KKT knowledge-transfer module not only helps to learn the tasks better (see Table \ref{tab:knn-results}) but also learn them faster. For an efficiency and robustness analysis of the KKT module, please refer to Appendix \ref{sec:app_kkt}.

\vspace{-3pt}
\subsection{Additional Results and Analysis}
\vspace{-3pt}

\begin{figure}[t!]
         \centering
        \includegraphics[width=0.75\linewidth]{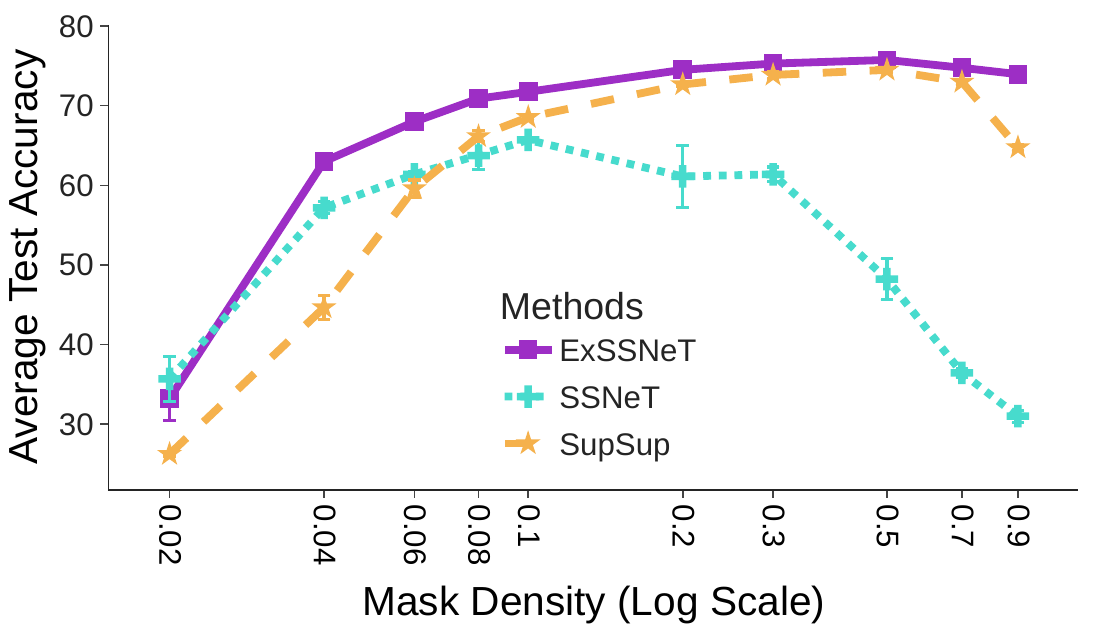}
        \vspace{-5pt}
        \captionsetup{margin=0.25cm}
        \captionof{figure}{\label{fig:sparsity}Average test accuracy versus mask density on SplitCIFAR100 dataset.}
        
\end{figure}

\begin{table}[t!]
        \centering
        \scriptsize
        \vspace{-5pt}
        \resizebox{0.9\linewidth}{!}{  
        \begin{tabular}{ccc}
                \toprule
                \textbf{Method} & \textbf{S-TinyImageNet} & \textbf{Avg. Sparse Overlap} \\
                \midrule
                \textbf{SupSup} & 90.34 (0.0) & 90.1 \\
                \textbf{\modelorg{}} & 89.02 (2.2) & 90.0 \\
                \textbf{\method{}} & \textbf{91.21 (0.0)} & 90.0 \\
                \bottomrule
            \end{tabular}
        }
        \vspace{-5pt}
        \captionsetup{margin=0.25cm}
        \captionof{table}{\label{tab:large-exp}
        Average accuracy $\uparrow$ \textcolor{BrickRed}{(\textbf{forgetting metric} $\downarrow$)} and average sparse overlap when learning 100 tasks.}
        \vspace{-15pt}
\end{table}

\paragraph{Q4. Effect of Mask Density on Performance:} Next, we show the advantage of using \method{} when the mask density is low. In Figure \ref{fig:sparsity}, we show the average accuracy for the Split-CIFAR100 dataset as a function of mask density. We observe that \method{} obtains $7.9$\%, $18.4$\%, $8.4$\%, and $4.7$\% improvement over SupSup for mask density values $0.02, 0.04, 0.06, 0.08$ respectively.
This is an appealing property as tasks select fewer parameters which inherently reduces sparse overlap allowing \method{} to learn a large number of tasks.

\looseness=-1
\paragraph{Q5. Can \method{} Learn a Large Number of Tasks?} 
SupSup showed that it can scale to a large number of tasks. Next, we perform experiments to learn 100 tasks created by splitting the TinyImageNet dataset. In Table \ref{tab:large-exp}, we show that this property is preserved by \method{} while resulting in a performance improvement over SupSup. We note that as the number of task increase, the sparse overlap between the masks also increases resulting in fewer trainable model weights. In the extreme case where there are no free weights, \method{} by design reduces to SupSup because there will be no weight training. Moreover, if we use larger models there are more free parameters, leading to even more improvement over SupSup.

\begin{table}[t!]
        
        \centering
        \resizebox{0.9\linewidth}{!}{  
        \begin{tabular}{clll}
        \toprule
        \textbf{Method} & \textbf{FastText} & \textbf{Glove} & \textbf{BERT}\\
        \midrule
        \textbf{SupSup} & 54.01 & 55.52 & 74.0 \\
        \textbf{\modelorg{}} & 60.41 [\textcolor{color2}{+6.4}] & 59.78 [\textcolor{color2}{+4.3}] & 74.5 [\textcolor{color2}{+0.5}] \\
        \textbf{\method{}} & \textbf{62.52} [\textcolor{color2}{+8.5}] & \textbf{62.81} [\textcolor{color2}{+7.3}] & \textbf{74.8} [\textcolor{color2}{+0.8}]\\
        \bottomrule
        \end{tabular}
        }
        \vspace{-5pt}
        \captionsetup{margin=0.25cm}
        \captionof{table}{\label{tab:text-ablation-init} Ablation result for token embeddings. We report average accuracy $\uparrow$ \textcolor{color2}{\textbf{[and gains over SupSup]}}}
        \vspace{-15pt}
\end{table}

\looseness=-1
\paragraph{Q6. Effect of Token Embedding Initialization for NLP:}
For our language experiments, we use a pretrained BERT model \citep{devlin-etal-2019-bert} to obtain the initial token representations. 
We perform ablations on the token embedding initialization to understand its impact on CL methods.
In Table \ref{tab:text-ablation-init}, we present results on the  S2\textsuperscript{\ref{foot:order}} task-order sequence of the sampled version of WebNLP dataset (see Section \ref{sec:setup}, Datasets). We initialize the token representations using \textit{FastText} \citep{bojanowski2016enriching-fasttext}, \textit{Glove} \citep{pennington-etal-2014-glove}, and \textit{BERT} embeddings.
From Table \ref{tab:text-ablation-init}, we observe that -- (1) 
the performance gap between \method{} and SupSup increases from $0.8\% \rightarrow 7.3\%$ and $0.8\% \rightarrow 8.5\%$ when moving from BERT to Glove and FastText initializations respectively. 
These gains imply that it is even more beneficial to use \method{} in absence of good initial representations, and (2) the performance trend, \method{} > \modelorg{} > SupSup is consistent across initialization.

\section{Related Work}
\label{sec:related}

\textbf{Regularization-based methods} estimate the importance of model components and add importance regularization terms to the loss function.
\citet{zenke2017continual} regularize based on the distance of weights from their initialization, whereas \citet{kirkpatrick2017overcoming,schwarz2018progress} use an approximation of the Fisher information matrix \citep{pascanu2013revisiting} to regularize the parameters.
In NLP, \citet{han2020continual,wang2019sentence} use regularization to constrain the relevant information from the huge amount of knowledge inside large language models (LLM).
\citet{huang-etal-2021-continual-idbr} first identifies hidden spaces that need to be updated versus retained via information disentanglement \citep{fu2017style,li2020complementary} and then regularize these hidden spaces separately. 

\textbf{Replay based methods} maintain a small memory buffer of data samples \citep{de2019continual,yan2022learning} or their relevant proxies \citep{rebuffi2017icarl} from the previous tasks and retrain on them later to prevent CF. \citet{chaudhry2018efficient} use the buffer during optimization to constrain parameter gradients. 
\citet{shin2017continual,kemker2018fearnet} uses a generative model to sample and replay pseudo-data during training, whereas
\citet{rebuffi2017icarl} replay distilled knowledge from the past tasks. 
\citet{d2019episodic} employ episodic memory along with local adaptation, whereas \citet{sun2019lamol} trains a language model to generate a pseudo-sample for replay. 

\textbf{Architecture based methods}can be divided into two categories: (1) methods that add new modules over time \citep{li2019learn,veniat2021efficient,douillard2021dytox}; and (2) methods that isolate the network's parameters for different tasks \citep{serra2018overcoming,Fernando2017pathnet,mallya2018packnet,Fernando2017pathnet}. \citet{rusu2016progressive} introduces a new network for each task while \citet{schwarz2018progress} distilled the new network after each task into the original one. 
Recent prompt learning-based CL models for vision \citep{wang2022dualprompt,wang2022l2p} assume access to a pre-trained model to learn a set of prompts that can potentially be shared across tasks to perform CL this is orthogonal to our method that trains from scratch.
\citet{mallya2018packnet} allocates parameters to specific tasks and then trains them in isolation which limits the number of tasks that can be learned. 
In contrast, \citet{mallya2018piggyback} use a frozen pretrained model and learns a new mask for each task but a pretrained model is crucial for their method's good performance. \citet{supsup} removes the pretrained model dependence and learns a mask for each task over a fixed randomly initialized network.
\method{} avoids 
the shortcomings of \citet{mallya2018packnet,mallya2018piggyback} and performs supermask subnetwork training to increase the representational capacity compared to \citep{supsup} while performing knowledge transfer and avoiding CF.

\section{Conclusion}
\label{sec:conclusion}

\looseness=-1
We introduced a novel Continual Learning method, \method{} (Exclusive Supermask SubNetwork Training), that delivers enhanced performance by utilizing exclusive, non-overlapping subnetwork weight training, overcoming the representational limitations of the prior SupSup method. Through the avoidance of conflicting weight updates, \method{} not only improves performance but also eliminates forgetting, striking a delicate balance. Moreover, the inclusion of the Knowledge Transfer (KKT) module propels the learning process, utilizing previously acquired knowledge to expedite and enhance the learning of new tasks. The efficacy of \method{} is substantiated by its superior performance in both NLP and Vision domains, its particular proficiency for sparse masks, and its scalability up to a hundred tasks.

\section*{Limitations}

Firstly, we note that as the density of the mask increases, the performance improvement over the SupSup method begins to decrease. This is due to the fact that denser subnetworks result in higher levels of sparse overlap, leaving fewer free parameters for new tasks to update. However, it is worth noting that even in situations where mask densities are higher, all model weights are still trained by some task, improving performance on those tasks and making our proposed method an upper bound to the performance of SupSup. Additionally, the model size and capacity can be increased to counterbalance the effect of higher mask density. Moreover, in general, a sparse mask is preferred for most applications due to its efficiency.

Secondly, we have focused on the task incremental setting of continual learning for two main reasons: (1) in the domain of natural language processing, task identities are typically easy to obtain, and popular methods such as prompting and adaptors assume access to task identities. (2) the primary focus of our work is to improve the performance of supermasks for continual learning and to develop a more effective mechanism for reusing learned knowledge, which is orthogonal to the question of whether task identities are provided during test time.

Moreover, it is worth noting that, similar to the SupSup method, our proposed method can also be extended to situations where task identities are not provided during inference. The SupSup paper presents a method for doing this by minimizing entropy to select the best mask during inference, and this can also be directly applied to our proposed method, ExSSNeT, in situations where task identities are not provided during inference. This is orthogonal to the main questions of our study, however, we perform some experiments on Class Incremental Learning in the appendix~\ref{sec:app_cil}. 

\section*{Acknowledgements}
We thank Marc'Aurelio Ranzato for the helpful discussions to formulate the initial idea. We thank the reviewers and Xiang Zhou, Swarnadeep Saha, and Archiki Prasad for their valuable feedback on this paper. 
This work was supported by NSF-CAREER Award 1846185, DARPA Machine-Commonsense (MCS) Grant N66001-19-2-4031, Microsoft Investigator Fellowship, and Google and AWS cloud compute awards. The views contained in this article are those of the authors and not of the funding agency.

\bibliography{acl2023}
\bibliographystyle{acl_natbib}

\appendix
\section{Appendix for \method{}}
\label{sec:appendix}

\subsection{Sparse Overlap to Quantify Parameter Interference}
\label{sec:app_spoverlap}
Next, we propose a measure to quantify parameter interference for a task $i$, i.e., the fraction of the parameters in mask $M_i$ that are already updated by some previous task. We define \textit{sparse overlap} as the difference between the number of parameters selected by mask $M_i$ and $M^{free}_i$ divided by the total parameters selected by $M_i$.
Formally, we define \textit{sparse overlap} (SO) between current supermask $M_i$ and supermasks for previous tasks $\{M_j\}_{j=1}^{i-1}$ as,
\begin{align}
    \centering
    \footnotesize
    \text{SO}(M_i, \{M_j\}_{j=1}^{i-1}) &= \frac{sum(M_i) - sum(M^{free}_i)}{sum(M_i)} \label{eqn:sp_overlap} \\
    \text{and} ~ M^{free}_i &= M_i \land \lnot (\lor_{j=1}^{i-1} (M_{j})) \nonumber
\end{align}

\noindent where $\land, \lor, \lnot$ are logical \textit{and}, \textit{or}, and \textit{not} operations. 

\begin{table}[t!]
    \resizebox{0.95\linewidth}{!}{  
    \begin{tabular}{cl}
    \toprule
    \textbf{ID} &  \textbf{Task Sequence} \\
    \midrule
    
    S1 & mnli \Arrow{.2cm} qqp \Arrow{.2cm} qnli \Arrow{.2cm} sst2 \Arrow{.2cm} cola  (Dec. data Size) \\
    \midrule
    S2 & ag \Arrow{.2cm} yelp \Arrow{.2cm} amazon \Arrow{.2cm} yahoo \Arrow{.2cm} dbpedia \\
    S3 & yelp \Arrow{.2cm} yahoo \Arrow{.2cm} amazon \Arrow{.2cm} dbpedia \Arrow{.2cm} ag \\
    S4 & dbpedia \Arrow{.2cm} yahoo \Arrow{.2cm} ag \Arrow{.2cm} amazon \Arrow{.2cm} yelp \\
    S5 & yelp \Arrow{.2cm} ag \Arrow{.2cm} dbpedia \Arrow{.2cm} amazon \Arrow{.2cm} yahoo\\  
    S6 & ag \Arrow{.2cm} yelp \Arrow{.2cm} yahoo \\
    S7 & yelp \Arrow{.2cm} yahoo \Arrow{.2cm} ag \\
    S8 & yahoo \Arrow{.2cm} ag \Arrow{.2cm} yelp \\
    
    \bottomrule
    \end{tabular}
    }
    \captionsetup{margin=0.25cm}
    \captionof{table}{\label{tab:task-order}Task sequences used in text experiments. For the GLUE dataset, we use order corresponding to decreasing train data size. Sequence S2-S8 are from \citep{huang-etal-2021-continual-idbr,d2019episodic,sun2019lamol}.}
    
\end{table}

\subsection{Space, Time, and Memory Complexity of \method{}}
\looseness=-1
For training, we store an additional set of scores on GPU with size as the model weight. The additional GPU memory required is a small fraction because the model activations account for a huge fraction of the total GPU memory. Our runtime is similar to training the weight of a model with $<5\%$ overhead due to the logical operations on masks and masking weight during the forward passes. For training time comparisons refer to Appendix Table \ref{app:tab:runtime}. On the disk, we need to store $k*|W|$ updated weights of 32-bits and boolean mask which takes 1-bit for each parameter. Hence, we take  $max(|W|*k*t, |W|) * 32 + |W| * 1$ bits in total as in the worst case we need to store all $|W|$ model weights.

\begin{table}[t!]
    \centering
    \resizebox{0.95\linewidth}{!}{  
    \begin{tabular}{cccccc}
    \toprule
    \textbf{Dataset}  & \textbf{Class} & \textbf{Type} & \textbf{Train}& \textbf{Validation}& \textbf{Test} \\
    \midrule
    AGNews  & 4 & News & 8k & 8k & 7.6k  \\
    Yelp & 5 & Sentiment& 10k & 10k & 7.6k \\
    Amazon & 5 & Sentiment & 10k &  10k & 7.6k \\
    DBPedia & 14 & Wikipedia & 28k &  28k & 7.6k \\
    Yahoo & 10 & Q\&A & 20k &  20k & 7.6k \\
    \bottomrule
    \end{tabular}
    }
    \caption{\label{app:tab:sampled-stats}
    Statistics for sampled data used from \citet{huang-etal-2021-continual-idbr} for hyperparameter tuning. The validation set is the same size as the train set. Class means the number of output classes for the text classification task. Type is the domain of text classification.
    }
    
\end{table}

\subsection{Experimental setup and hyperparameters}
\label{sec:additional-details}

Unless otherwise specified, we obtain supermasks with a mask density of 0.1. In our CNN models, we use non-affine batch normalization to avoid storing their means and variance parameters for all tasks \citep{supsup}. Similar to \citep{supsup}, bias terms in our model are 0 and we randomly initialize the model parameters using \textit{signed kaiming constant} \citep{ramanujan2019s}.
We use Adam optimizer \citep{kingma2014adam} along with cosine decay \citep{cosine} and conduct our experiments on GPUs with 12GB of memory. We used approximately 6 days of GPU runtime. For our main experiment, we run three independent runs for each experiment and report the averages for all the metrics and experiments. For natural language tasks, unless specified otherwise we initialize the token embedding for our methods using a frozen BERT-base-uncased \citep{devlin2018bert} model's representations using Huggingface \citep{wolf2020transformers}. We use a static CNN model from \citet{kim-2014-convolutional} as our text classifier over BERT representations. The model employs 1D convolutions along with \textit{Tanh} activation. The total model parameters are $\sim$110M Following \citet{sun2019lamol,huang-etal-2021-continual-idbr}, we evaluate our model on various task sequences as provided in Appendix Table \ref{tab:task-order}, while limiting the maximum number of tokens to 256. Following \citep{supsup}, we use LeNet \citep{lecun1998lenet} for SplitMNIST dataset, a Resnet-18 model with fewer channels \citep{supsup} for SplitCIFAR100 dataset, a ResNet50 model \citep{He2016resnet} for TinyImageNet dataset. Unless specified, we randomly split all the vision datasets to obtain five tasks with disjoint classes. We use the codebase of DER \citep{buzzega2020der} to obtain the vision baselines. In all our experiments, all methods perform an equal number of epochs over the datasets. We use the hyperparameters from \citet{supsup} for our vision experiments.

For the ablation experiment on natural language data, following \citet{huang-etal-2021-continual-idbr}, we use a sampled version of the WebNLP datasets due to limited resources. The reduced dataset contains 2000 training and validation examples from each output class. The test set is the same as the main experiments. The dataset statistics are summarized in Table \ref{app:tab:sampled-stats}. For WebNLP datasets, we tune the learning rate on the validation set across the values $\{0.01, 0.001, 0.0001\}$, for GLUE datasets we use the default learning rate of the BERT model. For our vision experiments, we use the default learning rate for the dataset provided in their original implementation. For TinyImageNet, SplitCIFAR100, SplitMNIST dataset, we run for 30, 100, and 30 epochs respectively. We store 0.1\% of our vision datasets for replay while for our language experiments we use 0.01\% of the data because of the large number of datasets available for them.

\begin{table}[t!]
    \centering
    \resizebox{0.8\linewidth}{!}{  
    \begin{tabular}{ccc}
        \toprule
        \textbf{Method} & \textbf{Average Accuracy} & \textbf{Forgetting} \\
        \midrule
        \textbf{SupSup} & 68.07 & 0.00 \\
        \textbf{ExSSNeT} & 74.77 & 0.00 \\
        \bottomrule
    \end{tabular}
    }
    \caption{\label{tab:imagenet} Comparision between \method{} and the best baseline SupSup on Imagenet Dataset.}
\end{table}

\subsection{Additional Results}

\subsubsection{Results on Imagenet Dataset}
\label{sec:app_imagenet}
In this experiment, we take the ImageNet dataset \cite{imagenet} with 1000 classes and divide it into 10 tasks where each task is a 100-way classification problem. In Table \ref{tab:imagenet}, we report the results for ExSSNeT and the strongest vision baseline method, SupSup. We omit other methods due to resource constraints. We observe a strong improvement of 6.7\% of \method{} over SupSup, indicating that the improvements of our methods exist for large scale datasets as well. 

\begin{table}[t!]
    \centering
    \resizebox{0.8\linewidth}{!}{  
    \begin{tabular}{cccccc}
        \toprule
        \textbf{K} & \textbf{1} & \textbf{5} & \textbf{10} & \textbf{20} & \textbf{50} \\
        \midrule
        \textbf{\method{}} & 71.38 & 71.66 & 71.01 & 70.46 & 69.74 \\
        \bottomrule
    \end{tabular}
    }
    \caption{\label{tab:vary_k} Effect of varying $k$ while keeping the number of batches used for the KKT module fixed.}
\end{table}

\begin{table}[t!]
    \centering
    \resizebox{0.8\linewidth}{!}{  
    \begin{tabular}{cccccc}
        \toprule
        \textbf{Num. Batches} & \textbf{2} & \textbf{5} & \textbf{10} & \textbf{50} & \textbf{100} \\
        \midrule
        \textbf{\method{}} & 70.65 & 70.63 & 71.01 & 71.07 & 71.6 \\
        \bottomrule
    \end{tabular}
    }
    \caption{\label{tab:vary_batch} Effect of varying the number of batches while keeping the $k$ for top-$k$ neighbours fixed for KKT module fixed.}
\end{table}

\begin{table}[t!]

\centering
    
    \resizebox{0.9\linewidth}{!}{  
    \begin{tabular}{cccc}
        \toprule
        \textbf{Method} & \textbf{S-MNIST} &  \textbf{S-CIFAR100} & \textbf{S-TinyImageNet} \\
        \midrule
        \textbf{SupSup} & 22.6 & 18.9 & 18.1 \\
        \textbf{+ KKT} & 46.4 & 48.3 & 52.4 \\
        \midrule
        \textbf{\modelorg{}} & 22.5 & 17.6 & 18.6 \\
        \textbf{+ KKT} & 52.7 & 49.9 & 52.4\\
        \midrule
        \textbf{\method{}} & 22.5 & 17.3 & 18.5 \\
        \textbf{+ KKT} & 47.8 & 48.8 & 52.4 \\
        \bottomrule
    \end{tabular}
    }
    \captionsetup{margin=0.25cm}
    \captionof{table}{\label{tab:overlap} We report the \textcolor{BrickRed}{average sparse overlap} for all method and dataset combinations reported in Table \ref{tab:knn-results}.}
\end{table}

\begin{figure}[t!]
\centering
\includegraphics[width=0.9\linewidth]{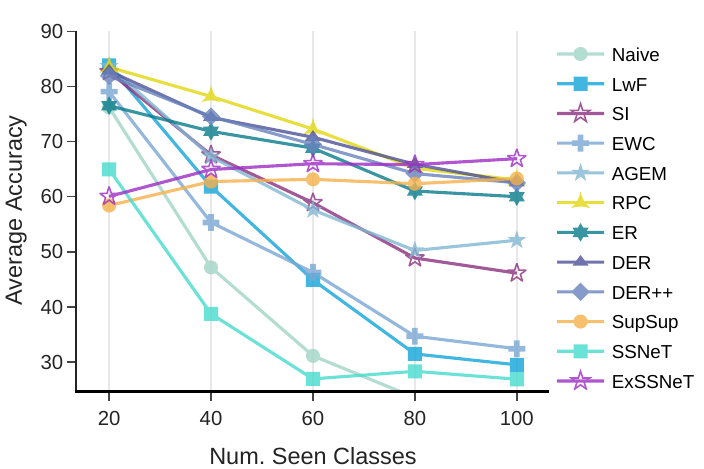}
\vspace{5pt}
\captionof{figure}{\label{fig:acc_evolution}Average Accuracy of all seen tasks as a function of the number of learned classes for the Split-CIFAR100 dataset.}
\end{figure}

\subsubsection{Analysis of Efficiency, Runtime, and hyperparameters of the KKT module}
\label{sec:app_kkt}
Firstly, we would like to note that the KKT module is lightweight and efficient because it only runs once for each task before we start training on it and only uses a few batches to estimate the relevant mask. Given that we perform multiple epochs over the task’s data, the cost of the KKT module becomes negligible in comparison to it and runs in almost similar clock time as without it. The runtime on splitcifar100 datasets with 100 epochs for ExSSNeT is 168 minutes and for ExSSNeT + KKT is 173 minutes which is a very small difference.

Second, there are two main hyperparameters in the KKT module – (1) $k$ for taking the majority vote of top-$k$ neighbors, and (2) the total number of batches used from the current task in this learning and prediction process. We present additional results on the splitcifar100 dataset when changing these hyperparameters one at a time.

In Table \ref{tab:vary_k}, we use 10 batches for KKT with a batch size of 64, resulting in 640 samples from the current task used for estimation. We report the performance of \method{} when varying $k$. From this table, we observe that the performance increases with $k$ and then starts to decrease but in general most values of $k$ work well.

Next, in Table \ref{tab:vary_batch}, we use a fixed $k$=10 and vary the number of batches used for KKT with a batch size of 64 and report the performance of \method{}. We observe that as the number of batches used for finding the best mask increases the prediction accuracy increases because of better mask selection. Moreover, as few as 5-10 batches work reasonably well in terms of average accuracy.

From both of these experiments, we can observe that the KKT module is fairly robust to different values of these hyperparameters but carefully selecting them hyperparameters can lead to slight improvement.

\subsubsection{Class Incremental Learning}
\label{sec:app_cil}
We performed Class Incremental Learning experiments on the TinyImageNet dataset (10-tasks, 20-classes in each) and used the One-Shot algorithm from SupSup \cite{supsup} to select the mask for inference. Please refer to Section-3.3 and Equation-4 of the SupSup paper \cite{supsup} for details. 
From Table~\ref{tab:cil}, we observe that \method{} outperforms all baseline methods that do not use Experience Replay by at least 2.75\%. Moreover, even with the need for a replay buffer, \method{} outperforms most ER-based methods and is comparable to that of DER.

\begin{table}[t!]
    \centering
    \begin{tabular}{ccc}
        \toprule
        \textbf{Method} &	\textbf{BufferSize} &	\textbf{TinyImageNet} \\
        \midrule
        \textbf{SGD} &	0 &	7.92 \\
        \textbf{oEWC} &	0 &	7.58 \\
        \textbf{LwF} &	0 &	8.46 \\
        \textbf{ER} &	200 &	8.49 \\
        \textbf{A-GEM} &	200 &	8.07 \\
        \textbf{iCARL} &	200 &	7.53 \\
        \textbf{DER} &	200 &	11.87 \\
        \textbf{DER++} &	200 &	10.96 \\
        \midrule
        \textbf{SupSup} &	0 &	10.27 \\
        \textbf{ExSSNeT} &	0 &	11.21 \\
        \bottomrule
    \end{tabular}
    \caption{\label{tab:cil} Results for CIL setting.}
    
\end{table}

\subsubsection{Sparse Overlap Numbers}
In Table \ref{tab:overlap}, we report the sparse overlap numbers for SupSup, \modelorg{}, and \method{} with and without the KKT knowledge transfer module. This table corresponds to the results in main paper Table \ref{tab:knn-results}.

\subsubsection{Average Accuracy Evolution}
\label{sec:acc-evolution}

In Figure \ref{fig:acc_evolution}, we plot $\sum_{i \leq t} A_{ti}$ vs $t$, that is the average accuracy as a function of observed classes. This plot corresponds to the SplitCIFAR100 results provided in the main paper Table \ref{tab:vision-main}. We can observe from these results that Supsup and ExSSNeT performance does not degrade when we learn new tasks leading to a very stable curve whereas for other methods the performance degrades as we learn new tasks indicating some degree of forgetting.

\vspace{10pt}
\begin{algorithm}[H]
\scriptsize
\caption{\label{alg:training} \method{} training procedure.}
\DontPrintSemicolon
\KwIn{Tasks $\mathcal{T}$, a model
$\mathcal{M}$, mask sparsity $k$, exclusive=True}
\KwOut{Trained model}
\Comment{\scriptsize Initialize model weights $W^{(0)}$}
initialize\_model\_weights($\mathcal{M}$)

\ForAll{$i \in range(|\mathcal{T}|)$}
{
     \Comment{ \scriptsize Set the mask $M_i$ corresponding to task $t_i$ for optimization.}
    mask\_opt\_params = $M_i$ 
    
    \Comment{\scriptsize Learn the supermask $M_i$ using edge-popup}
    \ForAll{$em \in \text{mask\_epochs}$}
    {
        $M_i$ = learn\_supermask(model, mask\_opt\_params, $t_i$)
    }
    
    \Comment{\scriptsize Model weight at this point are same as the last iteration $W^{(i-1)}$}
    \uIf{$i > 1$ and exclusive}{
        \Comment{\scriptsize Find mask for all the weights used by previous tasks.}
        $M_{1:i-1} = \lor_{j=1}^{i-1} (M_{j})$
        
        \Comment{\scriptsize Get mask for weights in $M_i$ which are not in $\{M_i\}_{j=1}^{i-1}$}
        $M_i^{free} = M_i \land \lnot M_{1:i-1}$
        
        \Comment{\scriptsize Find non-overlapping weight for updating.}
        $W^{(i)}_{free} = M_i^{free} \odot W^{(i-1)}$
    }\;
    
    \ElseIf{not exclusive}{
    $W^{(i)}_{free} = W^{(i-1)}$ 
    }
    weight\_opt\_params = $W^{(i)}_{free}$
    
    \Comment{\scriptsize Learn the free weight in the supermask $M_i$}
    \ForAll{$em \in \text{weight\_epochs}$}
    {
        $W^{(i)}$ = update\_weights(model, weight\_opt\_params, $t_i$)
    }
}
\end{algorithm}

\subsubsection{Runtime Comparison across methods}
\label{sec:runtime}

In this Section, we provide the result to compare the runtime of various methods used in the paper. We ran each method on the sampled version of the WebNLP dataset for the \textit{S2} task order as defined in Table \ref{tab:task-order}. We report the runtime of methods for four epochs over each dataset in Table \ref{app:tab:runtime}. Note that the masking-based method, SupSup, \modelorg{}, \method{} takes much lower time because they are not updating the BERT parameters and are just finding a mask over a much smaller CNN-based classification model using pretrained representation from BERT. This gives our method an inherent advantage that we are able to improve performance but with significantly lower runtime while learning a mask over much fewer parameters for the natural language setting.

\begin{table}[t!]
    \centering
    \resizebox{0.8\linewidth}{!}{  
    \begin{tabular}{cc}
    \toprule
    \textbf{Method} & \textbf{Runtime (in minutes)} \\
    \midrule
    \textbf{\textit{Multitask}} & 200 \\
    \textbf{Finetune} & 175 \\
    \textbf{Replay} & 204 \\
    \textbf{AdapterBERT + FT} & 170 \\
    \textbf{AdapterBERT + Replay} & 173 \\
    \textbf{MultiAdaptBERT} & 170 \\
    \textbf{Regularization} &  257 \\
    \textbf{IDBR} & 258 \\
    \textbf{SupSup} & \textbf{117} \\
    \midrule
    \textbf{\modelorg{}} & \textbf{117} \\
    \textbf{\method{}} & \textbf{117} \\
    \bottomrule
    \end{tabular}
    }
    \caption{\label{app:tab:runtime}Runtime comparison of different methods used in the text experiments.}
    
\end{table}

\begin{table}[t!]
    \centering
    \footnotesize
    \vspace{5pt}
    \resizebox{0.9\linewidth}{!}{  
    \begin{tabular}{llllllllll}
    \toprule
    \textbf{Method ($\downarrow$)} & \multicolumn{1}{c}{\textbf{GLUE}}  & \multicolumn{5}{c}{\textbf{WebNLP}} \\
    \cmidrule(lr){2-2} \cmidrule(lr){3-7}
    \textbf{Order ($\rightarrow$)}& \textbf{S1} & \textbf{S2} & \textbf{S3} & \textbf{S4} & \textbf{S5} & \textbf{Average}\\
    \midrule
    \textbf{\textit{Random}} & \textit{33.3} & \textit{7.14} & \textit{7.14} & \textit{7.14} & \textit{7.14} & \textit{7.14} \\
    \textbf{\textit{Multitask}} & \textit{80.6} & \textit{77.4} & \textit{77.5} & \textit{76.9} & \textit{76.8} & \textit{77.1} \\
    \midrule
    \textbf{FT} & 14.0 & 27.0 & 22.9 & 30.4 & 15.6 & 24.0 \\
    \textbf{Replay} & 79.7 & 75.2 & 74.5 & 75.2 & 75.5 & 75.1 \\
    \textbf{AdaptBERT + FT} & 25.1 & 20.8 & 19.1 & 23.6 & 14.6 & 19.5 \\
    \textbf{AdaptBERT + Replay} & 78.6 & 73.3 & 74.3 & 74.7 & 74.6 & 74.2 \\
    \textbf{MultiAdaptBERT} & 83.6 & 76.7 & 76.7 & 76.7 & 76.7 & 76.7\\
    \textbf{Regularization} & 75.5 & 75.9 & 75.0 & 76.5 & 76.3 & 75.9 \\
    \textbf{IDBR} & 77.5 & 75.8 & 75.4 & 76.4 & 76.4 & 76.0 \\
    \textbf{SupSup} & 78.1 & 75.7 & 76.0 & 76.0 & 75.9 & 75.9 \\
    \midrule
    \textbf{\modelorg{}} & 77.2 & 76.3 & 76.3 & 77.0 & 76.1 & 76.4 \\
    \textbf{\method{}} & 80.1 & 77.1 & 77.3 & 77.2 & 77.1 & 77.2 \\
    
    \bottomrule
    \end{tabular}
    }
    \captionof{table}{\label{app:tab:text-main-val} Average validation accuracy ($\uparrow$) for multiple tasks and sequence orders with previous state-of-the-art (SotA) methods.}
    
\end{table}

\subsubsection{Validation results} 
\label{sec:val-results}
In Table \ref{app:tab:text-main-val}, we provide the average validation accuracies for the main natural language results presented in Table \ref{tab:text-main}. We do not provide the validation results of LAMOL \citep{sun2019lamol} and MBPA++ \citep{d2019episodic} as we used the results provided in their original papers. For the vision domain, we did not use a validation set because no hyperparameter tuning was performed as we used the experimental setting and default parameters from the original source code from \citep{supsup,wen2020batchensemble}.

\begin{table*}[tbh!]
\centering
\large
\resizebox{0.8\linewidth}{!}{  
\begin{tabular}{ccccccccc}
\toprule
\textbf{Model ($\downarrow$)} &  \multicolumn{4}{c}{\textbf{Length-5 WebNLP}} & \multicolumn{4}{c}{\textbf{Length-3 WebNLP}}\\
\cmidrule(lr){2-5} \cmidrule(lr){6-9}
\textbf{Order ($\rightarrow$)} & \textbf{S2} & \textbf{S3} & \textbf{S4} & \textbf{Average}  & \textbf{S6} & \textbf{S7} & \textbf{S8} & \textbf{Average} \\
\midrule
\textbf{\textit{Random}} & \textit{7.14} & \textit{7.14} & \textit{7.14} & \textit{7.14} & \textit{10.0} & \textit{10.0} & \textit{10.0} & \textit{10.0} \\
\textbf{\textit{MTL}} & \textit{75.09} & \textit{75.09} & \textit{75.09} & \textit{75.09} & \textit{74.16} & \textit{74.16} & \textit{74.16} & \textit{74.16} \\
\midrule
\textbf{Finetune}$\dagger$   & 32.37 & 32.22 & 26.44 & 30.34 & 25.79 & 36.56 & 41.01 & 34.45  \\
\textbf{Replay}$\dagger$ & 68.25 & 70.52 & 70.24 & 69.67 & 69.32 & 70.25 & 71.31 & 70.29  \\
\textbf{Regularization}$\dagger$ & 72.28 & 73.03 & 72.92 & 72.74 & 71.50 & 70.88 & 72.93 & 71.77  \\
\textbf{AdaptBERT}  & 30.49 & 20.16 & 23.01 & 24.55 & 24.48 & 31.08 & 26.67 & 27.41  \\
\textbf{AdaptBERT + Replay} & 69.30 & 67.91 & 71.98 & 69.73 & 66.12 & 69.15 & 71.62 & 68.96  \\
\textbf{IDBR}$\dagger$   &  72.63 & 73.72 & 73.23 & 73.19 & 71.80 & 72.72 &  73.08 & 72.53  \\
\textbf{SupSup} & 74.01 & 74.04 & 74.18 & 74.08 & 72.01 & 72.35 & 72.53 & 72.29 \\ 
\midrule
\textbf{\modelorg{}} & 74.5 & 74.5 & 74.65 & 74.55 & 73.1 & 72.92 & 73.07 & 73.03 \\
\textbf{\method{}} & \textbf{74.78} & \textbf{74.72} & \textbf{74.71} & \textbf{74.73} & \textbf{72.67} & \textbf{72.99} & \textbf{73.24} & \textbf{72.97} \\
\bottomrule 
\end{tabular}
}
\captionsetup{margin=0.25cm}
\captionof{table}{\label{tab:text-seq-len}Average test accuracy reported over task sequences for three independent runs on sub-sampled data. Results with $\dagger$ are taken from \citet{huang-etal-2021-continual-idbr}.}
\end{table*}

\subsubsection{Effect of Task Order and Number of Tasks} 
\label{sec:task-order}

Following \citet{huang-etal-2021-continual-idbr}, we conduct experiments to study the effect of task length and order in the language domain. We use task sequences of lengths three and five, with multiple different task orders on the sampled data (Section \ref{sec:setup}, Table \ref{tab:task-order}, and Appendix) to characterize the impact of these variables on the performance. In Table \ref{tab:text-seq-len}, we present the average test accuracy averaged over three different random seeds. We observe that across all six different settings our method performs better compared to all the baseline methods. Our methods bridge the gap toward multitask methods' performance, leaving a gap of  0.36\% and 1.19\% for lengths three and five sequences, respectively.

\subsection{Additional Model Details}

\subsubsection{Algorithm for \method{}}
\label{sec:exessnet-algorithm}

In Algorithm \ref{alg:training}, we provide a pseudo-code for our method \method{} for easier reference and understanding. We also attach our working code as supplementary material to encourage reproducibility.

\subsubsection{Model Diagram for Supsup}
\label{sec:supsup}

\begin{figure*}[tbh]
    \centering
    \includegraphics[width=0.6\linewidth]{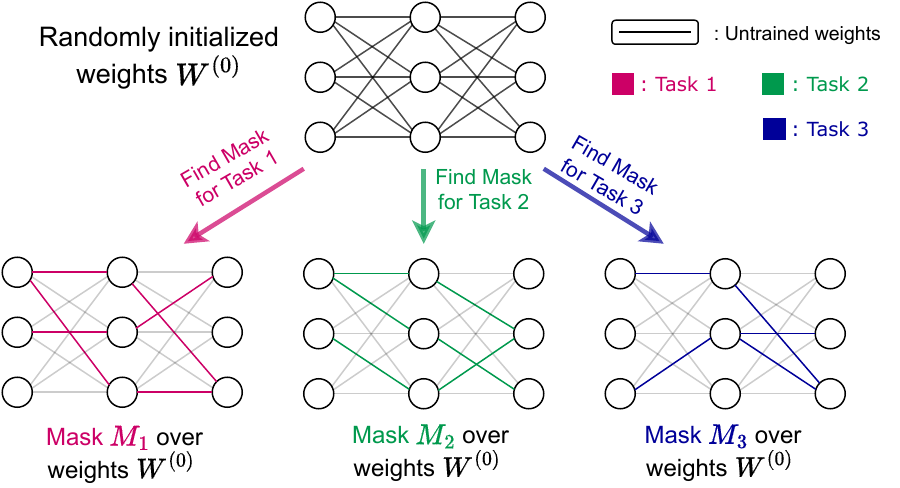}
    \captionof{figure}{This is a canonical model diagram for SupSup. In SupSup, the model weights are always fixed at the random initialization $W^{(0)}$. For each task SupSup learns a new mask (in this case $M_1, M_2, M_3$) over the weights $W^{(0)}$. A mask selectively activates a subset of weights for a particular task. This subset of selected weights forms a subnetwork inside the full model which we refer to as the supermask subnetwork. For example, when learning Task 2, SupSup learns the mask $M_2$ (the weights activated by the mask are highlighted in green) over the fixed weight $W^{(0)}$. These highlighted weights along with the participating nodes are the subnetwork formed by mask $M_2$. Whenever a prediction is made for Task 2 samples, this mask is selected and used to obtain the predictions. Please note that the model weights $W^{(0)}$ are never updated after their random initialization. Hence, for SupSup there is no learned knowledge sharing across tasks. This is in contrast to our setup in Figure \ref{fig:method}, where for the first task the mask is learned over the weights $W^{(0)}$ but once the mask is selected the weights of the corresponding subnetwork are also updates to obtain new weight $W^{(1)}$. Then the next task's mask is learned over these new set of weights $W^{(1)}$ and so on. Also note that in Figure \ref{fig:method}, we do not show the KKT knowledge transfer module here to avoid confusion.}
    \label{fig:supsup}
\end{figure*}

In Figure \ref{fig:supsup}, we provide the canonical model diagram for SupSup. Please read the figure description for more details regarding the distinctions between SupSup and ExSSNeT.

\end{document}